# A Neuronal Planar Modeling for Handwriting Signature based on Automatic Segmentation

I.Abroug Ben Abdelghani
Higher Institute of Applied Sciences and Technology of Kairouan, UR: SAGE-ENISo, University of Sousse, Tunisia.

N.Essoukri Ben Amara
National Engineering School of Sousse, UR: SAGE-ENISo, University of Sousse, Tunisia.

## ABSTRACT
This paper deals with offline handwriting signature verification. We propose a planar neuronal model of signature image. Planar models are generally based on delimiting homogenous zones of images; we propose in this paper an automatic segmentation approach into bands of signature images. Signature image is modeled by a planar neuronal model with horizontal secondary models and a vertical principal model. The proposed method has been tested on two databases. The first is the one we have collected; it includes 6000 signatures corresponding to 60 writers. The second is the public GPDS-300 database including 16200 signature corresponding to 300 persons. The achieved results are promising.

## 1. INTRODUCTION
Planar model or pseudo 2D model is an approach to model the horizontal and vertical variations of images in a simplified manner. This model offers the advantage of being treated as a nested 1D model rather than a truly 2D one. Therefore, it has the virtue of avoiding the insufficiency of 1D modelling as well as the complexity of 2D processing.

A planar model is basically a Hidden Markov Model (HMM) whose emission probabilities are also modelled by HMM. The planar Hidden Markov Model (PHMM) approach consists in dividing the image into several united parts (horizontal, vertical bands or different homogenous zones) and associating to each delimited part a 1D HMM that we shall call the secondary model. The principal model is defined according to the other direction, making correlation in the observation generated by secondary models. The states of the principal model are called super-states (figure 1)[6].

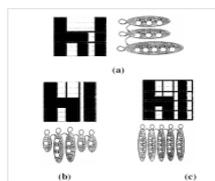

**Figure 1: Examples of PHMM architecture, (a): vertical principal model,(b, c): horizontal principal model, [4].**

PHMM has been applied in different fields such as the optical character recognition [4, 15], the recognition of handwritten digits [10], the Arabic off-line handwritten recognition [6], the recognition of printed and handwritten sub-words [6, 16, 17] and the classification of forms [21]. In all these planar architectures, each PHMM is based on an image segmentation step specific to the considered image. It is a primordial step whose goal is to divide the image into different homogenous parts depending on the morphology of the considered form.

Our paper deals with signature planar modeling in the context of its verification. In this work, we propose to extend the application field of planar models to different techniques such as Neural Networks (NN) [1, 2, 3]. The planar modeling of the handwritten signature needs a first phase of segmentation into homogeneous bands. However, the random variability of the signature form makes such segmentation very delicate. In fact, the handwritten signature does not follow any rule of morphology like handwriting.

Faced with this problem and being convinced that planar modelling of the signature image will be very beneficial for a system of handwritten signature verification, we have tried to resolve the problem of segmentation of the signature image in different ways. In a first study [3], each signature is delimited into three horizontal bands having the same height. The number of bands is randomly chosen since there is no physical criterion to highlight in the image. The achieved results with this approach are acceptable but the problem of delimiting homogenous bands remains not resolved. In a second study [1], we try a totally supervised segmentation; the signature image is divided into different horizontal bands manually. The segmentation criterion, the number of bands and the height of each band is the result of a morphology study of each class of the signature database. The achieved results are better than the first study [3], but this approach is totally overseen by the operator. In this paper, we propose to solve this problem with an automatic signature segmentation approach.

In the following section, we give an overview of the morphologic characteristics of the signature image and the different problems related to its verification. In the third section, we describe the system outline, especially the segmentation strategy, the feature extraction and the classification based on a planar model architecture. Experimentations and results are addressed in section 4. The conclusion of the work is presented in section 5.

## 2. THE MORPHOLOGIC CHARACTERISTICS OF SIGNATURE IMAGE

The handwritten signature is the result of a spontaneous and voluntary gesture realised by the individual's hand. Signing consists in affixing a personalised form of the handwriting to characterise a person uniquely. The signature can be represented by the surname, the name, and can be drawn graphically in a simple or complex way combining lines, graphics, points, etc. Many studies on the form of the handwritten signature have been carried out in order to accentuate its characteristics and the difficulties related to modelling this image [9, 18]. These studies mainly show that the handwritten signature depends on the physical and psychophysical state of the signer as well as the conditions in which the process is realised.

In our case, we have carried out a morphologic study of a great number of samples of handwritten signatures (Figure 2). For this, we have collected a set of 3600 signatures from 60 people having different origins (Tunisian, French), age, and social and cultural levels. Each person has provided 60 signatures taken at different instances.

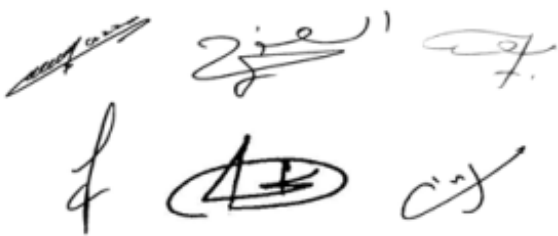

**Figure 2: Examples of studied signatures extracted from our database**

The literature and our study, has shown a great variability of the signature for the same person as well as for different people. In addition, we have raised the existence of forged signatures.

### 2.1 The interpersonal and intrapersonal variability

The signature is a random signal which presents various variations for different persons. These interpersonal variations depend on the environment at which the signature has been developed such as the origin, the cultural and individual level of the person (Figure 3). In fact, we can distinguish many classes of signatures in function of their origins: signatures of graphic forms (European origin), handwriting (American origin), cursive handwriting and graphic initials (Arabic origin), and the signatures of Asian origin which are easy to distinguish from other types of signatures [22].

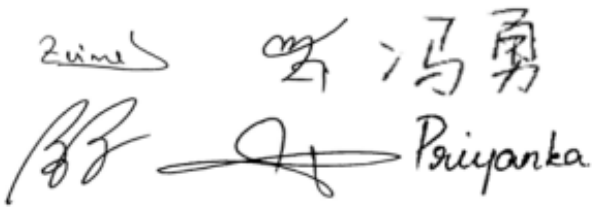

**Figure 3: Illustration of the interpersonal variability of the signature image**

Even for the same person, the signature presents many variations in the form. In fact, each signature sample is different from the previous signatures despite all the care we may give to reproduce a signature. It is the problem of interpersonal variability. These variations can be due to certain conditions like age, habits, moral state, practical conditions, etc. Figure 4 presents an example of intrapersonal variation of the signature.

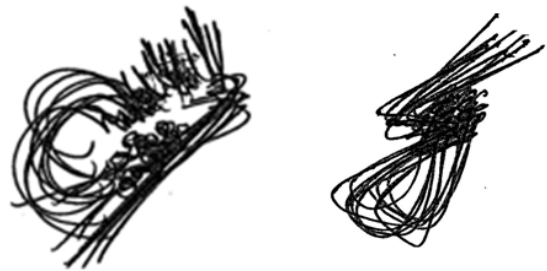

**Figure 4: The intrapersonal variability of the signature image**

### 2.2 The false signatures

The handwritten signature is particularly useful in verifying the identity, due to the fact that it has been part of daily transactions for a long time. Also, the handwritten signature is largely accepted by the great public to authentificate the official documents and give the individuals responsibilities face to involvements (contracts, financial and legal transactions, etc.) [9].

In addition, the signature is subject to falsifications. The class of forgeries is subdivided into many groups that have their distinctive signs. The groups of forgeries vary according to the forger, his ability and the effort he has contributed while producing the forgery and particularly whether he possesses a sample of the authentic signature or does not. The principal types of the less identical forgeries to the less easy original signatures to be distinguished are: the crude forgery, the random forgery, the simple forgery, the skilled forgery, the free-simulated forgery and the optically-transferred forgery [22]. According to the type of forgery, each group possesses its characteristics, which makes a study for each group of false signatures necessary before defining a system of handwritten signature verification [22].

All these inter/intrapersonal morphological variations linked to the handwritten signature, added to the falsifications, make modelling this very particular image an intricate task. Many systems have been developed in order to model and verify the image of the signature [5,7,8,11,12,14,19,20,23,24,25]. In our case, we have opted for planar modelling based on neural networks.

## 3. SYSTEM OUTLINE

Figure 5 shows the block diagram of the proposed system. The different signatures are normalized into [256 x 512] pixels. In the following sections, we describe the signature segmentation, the feature extraction and the classification steps. Classification is based on a planar model. Each person (class of signatures) is modelled by a planar model. The proposed planar architecture associates a secondary horizontal model for each band and a vertical principal model for presenting the correlation between different secondary models.

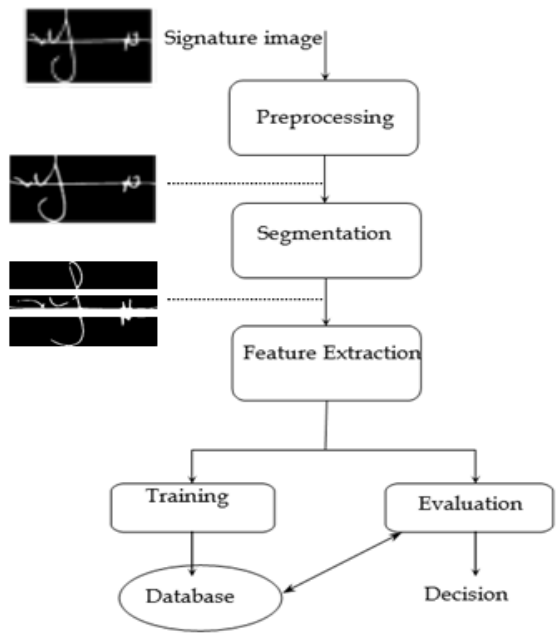

**Figure 5: The Block diagram of the system**

## 3.1 The proposed signature segmentation approach

The different studies we have done in the previous works of segmentation [1],[3] lead to a choice of 3 bands. In this work, the number of 3 bands has been confirmed by the segmentation approach we have adopted. The segmentation is based on the study of the distribution of vertical black pixel histogram. The goal is to collect consecutive lines having the same pixel distribution in the same band. Thresholding by the distribution means separates different lines of the image and creates bands. Figure 6 presents a sample of the signature image and its black pixel distribution. Thus, the successive lines whose proportion of the black pixels is more or equal, or less than the mean make up a band of homogeneous data. In the example illustrated in figure 6, the indexes i1, i2… i6 correspond to the line indexes that present bands.

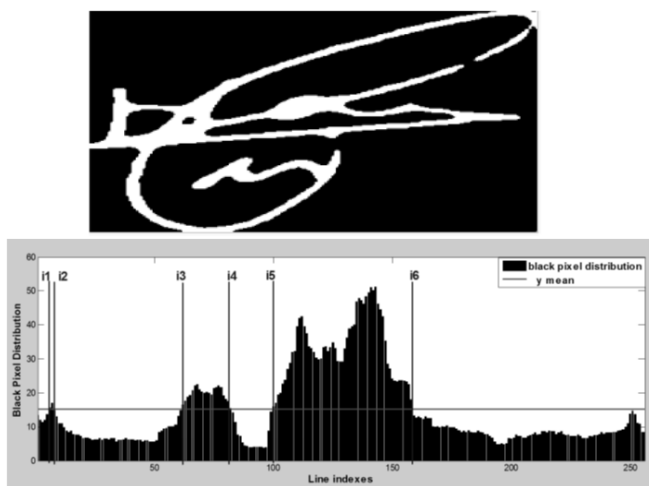

**Figure 6: Example of black pixel distribution of a signature image**

As shown in figure 6, the segmentation can sometimes generate a very low height bands, not having any physical signification (i1, i2, i3 and i4). Therefore, we have fixed a minimum threshold to get bands of significant data. The different realised experimentations have led to choosing a minimum height of 35 pixels. So, the result of the segmentation of the image in figure 6 corresponds to the indexes i2 and i4. Figure 7 gives the results of the segmentation step.

This strategy is applied for all the signature database classes. Each class is characterized by its intrapersonal variability; the heights of different bands are variable in the same class. This value is fed as a characteristic in the principal model (the vertical one).

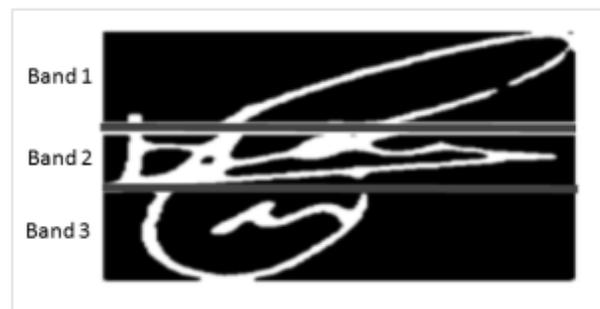

**Figure 7: Example of the result of signature segmentation**

## 3.2 Feature extraction

To reinforce the description of the signature variations, we retain two basic global feature types: geometric features and textural ones extracted from the application of the wavelet transform.

As a geometric feature, we select the orientation of the signature image, the number of black pixels in each band, the maximum number of black pixels corresponding to the vertical projection of the band and the height of each band.

As a textural feature, we apply the wavelet transform for each band separately. The obtained wavelet coefficients are retained as features for characterization. To reduce the number of features, we conduct a study on the pertinence of different extracted characteristics, and we retain the mean and the standard deviation of the approximation image and the standard deviation of the horizontal, vertical and diagonal details.

The obtained features are used as input vectors for the secondary and principal model. Each secondary model has a six-characteristic input vector (the wavelet features and the number of black pixels of each subband). The input vector of the principal model contains seven global attributes corresponding to the maximum number of black pixel of vertical projection lines, the height of each band and the horizontally-related signature orientation.

## 3.3 Classification

We use a one-class-one-network architecture; for each signature class we associate a planar model with a vertical principal model and three horizontal secondary models. Signature classes correspond to the different people representing the database.

Secondary and principal models are of multi-layer perceptron neural networks (MLP-NN) type. Each signature is modeled by three secondary MLP-NNs (one MLP-NN per band) and a principal MLP-NN. Each secondary model describes the morphologic variability of the associated band and decides whether it belongs to the involved signer. The role of each principal model is to make the global decision of the planar model.

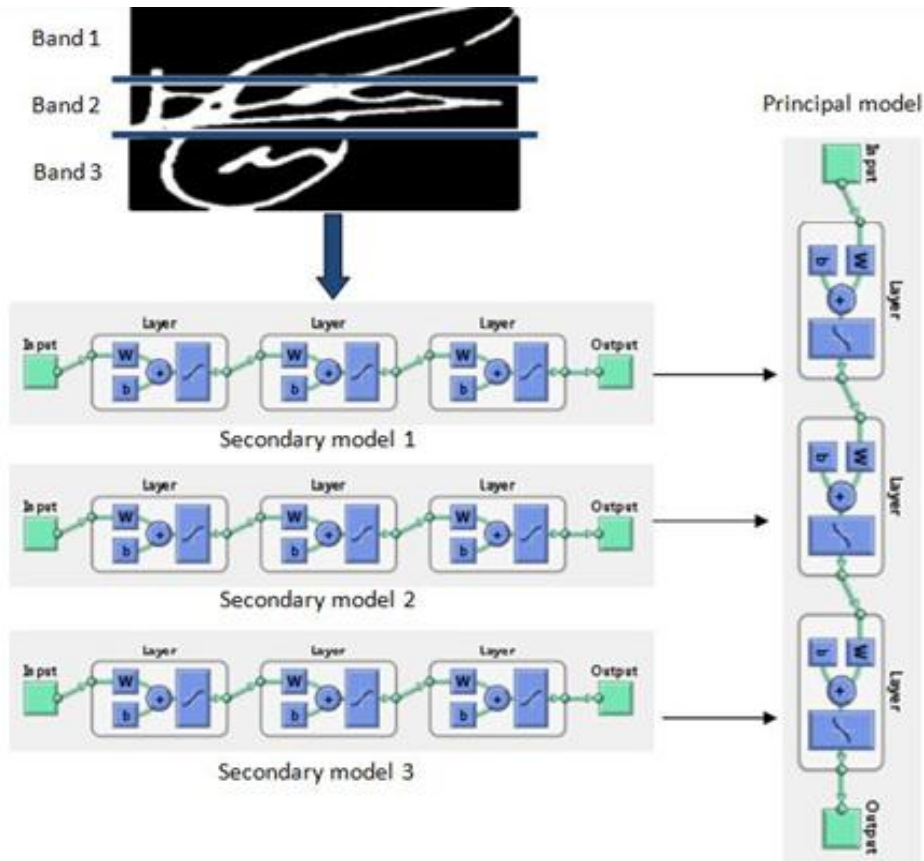

**Figure 8: Proposed planar neuronal model architecture**

## 4. EXPERIMENTATIONS AND RESULTS
In this section, we describe the used databases, different experimentations and the achieved results.

### 4.1 Database description
The experimental results are conducted on two signature databases.

The first is our proper database (database1) that contains 3600 genuine signatures corresponding to 60 peoples having different origins (Tunisian and French), age and cultural levels. 2400 forgery signatures (simple and skilled types) are added to the genuine samples. The forgeries are collected by 10 forgers (different from the genuine persons); each forger gives two simple forgeries and two skilled forgeries without any training.
Figure 9 gives examples of genuine and forgery signatures extracted from our database.

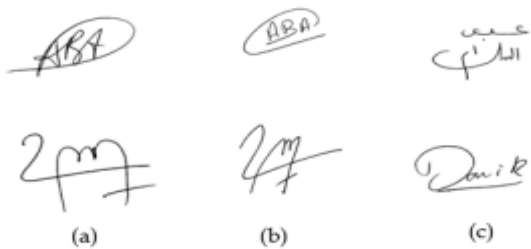

**Figure 9: Examples of the signature database; (a) genuine signatures, (b) skilled forgeries and (c) simple forgeries**

The second experimented database is the GPDS-300 database [24]. It contains data from 300 individuals of a Spanish origin: 24 genuine signatures for each individual, plus 30 forgeries of his signature. The 24 genuine specimens of each signer have been collected in a single day writing sessions. The forgeries have been produced from the static image of the genuine signature. Each forger has been allowed to practice the signature for as long as he wishes. Each forger imitated 3 signatures of 5 signers in a single day writing session. The genuine signatures shown to each forger are chosen randomly from the 24 genuine ones. Therefore for each genuine signature there are 30 skilled forgeries made by 10 forgers from 10 different genuine specimens, [24].

### 4.2 Experimentations
The proposed system has been tested on the two databases with the same experimental protocol.

Each signature image is segmented into 3 horizontal bands having different heights. Figure 10 gives the result of the segmentation step of two samples of signature images extracted from database-1.

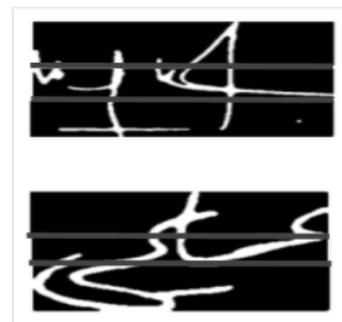

**Figure 10: Examples of two segmented signature images extracted from database-1**

Each segmented signature band is characterized by the textural and geometric features. For the wavelet features, we have tried a variety of scaling functions such as Haar basis, Daubechie's basis, Symlets basis, and the bi-orthogonal. The Symlets 6 with two levels of decomposition have led to the best results.

As far as the classification process is concerned, each created NN has three layers: first, the input layer contains a number of neurons corresponding to the size of the inputs feature vector. Second, the output layer contains a single neuron corresponding to the decision of the network. Third, the neuron number corresponding to the hidden layer is determined experimentally; it corresponds to the best NN performances.

In order to train and evaluate neural networks, we divide each signature database into training and evaluating data sets (Table 1 & Table 2).

**Table 1. Training and test data sets of database1**

| Training data set | 2400 genuine | |
|---|---|---|
| Evaluating data set | Random | 1200 |
| | Simple | 1200 |
| | Skilled | 1200 |

**Table 2. Training and test data sets of the GPDS-300 database**

| Training data set | 3600 genuine | |
|---|---|---|
| Evaluating data set | Random | 3600 |
| | Skilled | 9000 |

The training phase is carried out separately for each signature class in two stages with the feed-forward-back propagation algorithm. We start initially with the training of the three secondary models separately, and then we carry out the training of the principal model.

The training of the secondary models is carried out with an equal number of truths and false samples of signatures for each model. The number of hidden neurons by bands is obtained empirically.

The training of the principal model is carried out in a particular way. In each algorithm iteration, the outputs of the different secondary models, the minimum and the maximum heights of each band, the maximum number of pixel of vertical projections lines of the three bands and the angle of orientation of the signature are fed to the module of training of the principal model.

Training procedure was repeated 4 times with different training false subsets for the purpose of obtaining consistent results.

For testing the database1, random, simple and skilled forgeries were taken into account. In the case of GPDS-300 database, we test random and skilled forgeries. The performances of the system were evaluated by estimating the tow commonly used types of errors: the first is the False Reject Rate (FRR), which takes place when a true signature is rejected. The second is the False Acceptance Rate (FAR), that is when a forgery is accepted as true signature.

### 4.3 Results with database-1

In order to evaluate the proposed approach of segmentation and its impact on the signature verification system, we compare the achieved performances with results of our precedent works (evaluated on the same database) [1],[3]. Table 3 gives the corresponding error rates.

**Table 3. The achieved results of the proposed system and precedent systems tested on database-1**

| | | [1] | [3] | The proposed system |
|---|---|---|---|---|
| FRR % | | 15.67 | 10.16 | 16.42 |
| FAR % | Random | 3.57 | 6.05 | 6.50 |
| | Simple | 0.58 | 4.84 | 2.50 |
| | Skilled | 14.08 | 17.54 | 15.60 |

Results show that the performances of automatic segmentation approach are quite similar to results with manual segmentation.

### 4.4 Results with GPDS-300 database

In order to situate our system in the literature, we give in table 4 the results of our system with theirs of other systems evaluated on the GPDS-300 database.

**Table 4. The achieved results of the proposed system and other systems tested on GPDE-300 database**

| | | [12] | [20] | The proposed system |
|---|---|---|---|---|
| FRR % | | 5.64 | 25 | 14 |
| FAR % | Random | ---- | 22 | 0.92 |
| | Skilled | 5.37 | 26 | 40.4 |

We note that our results are suitable with random forgeries, but should be ameliorated with skilled ones.

### 5. CONCLUSION

In this paper we have proposed a neuronal planar model of the signature image based on an automatic segmentation approach. Signature images are delimited into three horizontal bands according to the distribution of black pixels in different pixel lines. The planar model architecture is based on three horizontal secondary models and a vertical principal one. Each band is characterized by geometric features and texture ones issued from wavelets. The proposed system is evaluated on two databases; our database and the public GPDS-300 database. The obtained results of the system are suitable. Different tests are underway to improve performance particularly in terms of skilled signatures.